%% file: main.tex
\pdfoutput=1

\documentclass[11pt]{article}

\usepackage[]{latex/acl}

\usepackage{times}
\usepackage{latexsym}

\usepackage[utf8]{inputenc} 
\usepackage[T1]{fontenc}    
\usepackage{microtype}      
\usepackage{xcolor}         
\usepackage{graphicx}
\usepackage{adjustbox}
\usepackage{enumerate}
\usepackage{makecell}
\usepackage{subcaption}
\usepackage{algorithm, algorithmic}
\usepackage{booktabs}
\usepackage{kotex}

\usepackage{amsmath}
\usepackage{amssymb}
\usepackage{bbm, dsfont}
\usepackage{mathtools}
\usepackage{amsthm}
\usepackage[normalem]{ulem}
\usepackage{xcolor,colortbl}
\usepackage{multirow}
\usepackage{cutwin}
\usepackage{arydshln}
\usepackage{caption}
\usepackage{enumitem}
\usepackage{wrapfig}
\usepackage{microtype}
\usepackage{graphicx}
\usepackage{bm}
\usepackage{eqparbox}
\usepackage{makecell}
\usepackage{threeparttable}
\usepackage{xspace}
\usepackage{tcolorbox}
\usepackage{pifont}
\usepackage{url}
\usepackage{tikz}
\usepackage{comment}
\usepackage{amsfonts}       
\usepackage{color}
\usepackage[nameinlink]{cleveref}
\crefname{section}{Sec.}{Secs.}
\Crefname{section}{Section}{Sections}
\crefformat{section}{\S#2#1#3}

\crefname{table}{Tab.}{Tabs.}
\Crefname{figure}{Fig.}{Figs.}

\newcommand{\ch}{\textsc{Changed}\xspace}
\newcommand{\unch}{\textsc{Unchanged}\xspace}
\newcommand{\new}{\textsc{New}\xspace}
\newcommand{\edited}{\textsc{Edited}\xspace}
\newcommand{\out}{\textsc{Outdated}\xspace}
\newcommand{\up}{\textsc{Updated}\xspace}

\newcommand{\cmark}{\textcolor{darkgreen}{\ding{51}}}
\newcommand{\xmark}{\ding{55}}

\newcommand{\eat}[1]{}

\definecolor{Red}{rgb}{0.6,0,0}
\definecolor{Blue}{rgb}{0,0,0.8}
\definecolor{Green}{rgb}{0,0.4,0.7}
\definecolor{airforceblue}{rgb}{0.36, 0.54, 0.66}
\definecolor{ao(english)}{rgb}{0.0, 0.5, 0.0}
\definecolor{azure(colorwheel)}{rgb}{0.0, 0.5, 1.0}
\definecolor{crimson}{rgb}{0.86, 0.08, 0.24}
\definecolor{darkcerulean}{rgb}{0.03, 0.27, 0.49}
\definecolor{cobalt}{rgb}{0.0, 0.28, 0.67}
\definecolor{rosegold}{rgb}{0.72, 0.43, 0.47}
\definecolor{orange-red}{rgb}{1.0, 0.27, 0.0}
\definecolor{mountainmeadow}{rgb}{0.19, 0.73, 0.56}
\definecolor{malachite}{rgb}{0.04, 0.85, 0.32}
\definecolor{darkblue}{rgb}{0.0, 0.0, 0.55}
\definecolor{customblue}{rgb}{0.2, 0.35, 0.8}
\definecolor{gg}{gray}{0.92}
\newcolumntype{a}{>{\columncolor{gg}}c}
\definecolor{darkgreen}{RGB}{0, 170, 0}

\usepackage{inconsolata}

\title{Carpe Diem\thanks{
\textit{Carpe diem} is a Latin phrase that translates to ``Live in the present" in English. It encourages individuals to make the most of the present\includegraphics[width=0.03\linewidth]{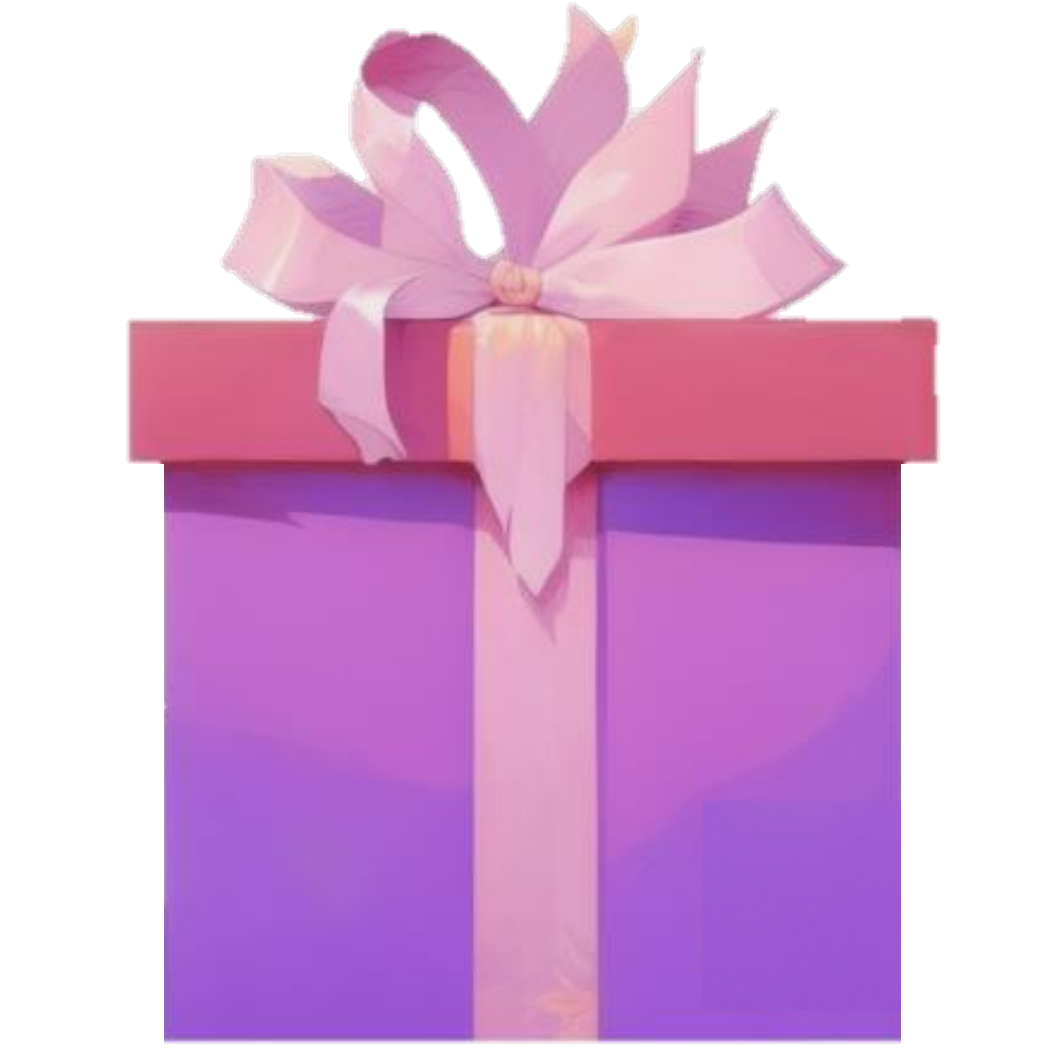} moment. \\
${\,\,\,\,\,\,\,\,}^{\dagger}$Corresponding authors}\includegraphics[width=0.04\linewidth]{figure/gift.pdf}: On the Evaluation of World Knowledge in\\Lifelong Language Models}

\author{Yujin Kim${}^{1}$ \quad Jaehong Yoon${}^{2}$ \quad Seonghyeon Ye${}^{1}$ \quad \bf Sangmin Bae${}^{1}$ \quad Namgyu Ho${}^{1}$ \\ \bf \quad Sung Ju Hwang${}^{1\dagger}$ \quad Se-young Yun${}^{1\dagger}$ \\
${}^{1}$KAIST AI \quad ${}^{2}$UNC Chapel Hill\\
\texttt{\{yujin399, sjhwang82, yunseyoung\}@kaist.ac.kr} }

\begin{document}
\maketitle
\input{tex/01_abstract}

\input{tex/02_intro}
\input{tex/04_dataset}
\input{tex/05_experiment}
\input{tex/06_discussion}
\input{tex/03_related}
\input{tex/07_conclusion}
\input{tex/09_limitation}
\input{tex/08_ethics}
\input{tex/10_ack}
\bibliography{reference}
\appendix
\input{tex/appx_01}
\input{tex/appx_02_additional_exp}


\end{document}

%% file: tex/01_abstract.tex
\begin{abstract}
The dynamic nature of knowledge in an ever-changing world presents challenges for language models trained on static data; the model in the real world often requires not only acquiring new knowledge but also overwriting outdated information into updated ones.
To study the ability of language models for these time-dependent dynamics in human language, we introduce a novel task, \textit{EvolvingQA}, a temporally evolving question-answering benchmark designed for training and evaluating LMs on an evolving Wikipedia database.
The construction of \textit{EvolvingQA} is automated with our pipeline using large language models.
We uncover that existing continual learning baselines suffer from updating and removing outdated knowledge.
Our analysis suggests that models fail to rectify knowledge due to small weight gradients.
In addition, we elucidate that language models particularly struggle to reflect the change of numerical or temporal information. 
Our work aims to model the dynamic nature of real-world information, suggesting faithful evaluations of the evolution-adaptability of language models. 
Our data construction code and dataset files are available at \url{https://github.com/kimyuji/EvolvingQA_benchmark}.
\end{abstract}

%% file: tex/02_intro.tex
\section{Introduction}
Large language models (LLMs)~\citep{radford2018improving, brown2020language, chowdhery2022palm, touvron2023llama} have demonstrated remarkable capabilities in encoding vast amounts of knowledge in massive training data, which can be applied for downstream tasks such as knowledge-intensive question-answering and multi-hop reasoning. However, knowledge is not static: scientific discoveries, cultural trends, and linguistic creativity are constantly updated and edited as the world changes. 
Current LLMs are trained on static data, implying that the encoded knowledge could go wrong as time passes, which affects their reasoning abilities \citep{dhingra2022time}.

\begin{figure}[t!]
    \centering
    
    \includegraphics[width=\linewidth]{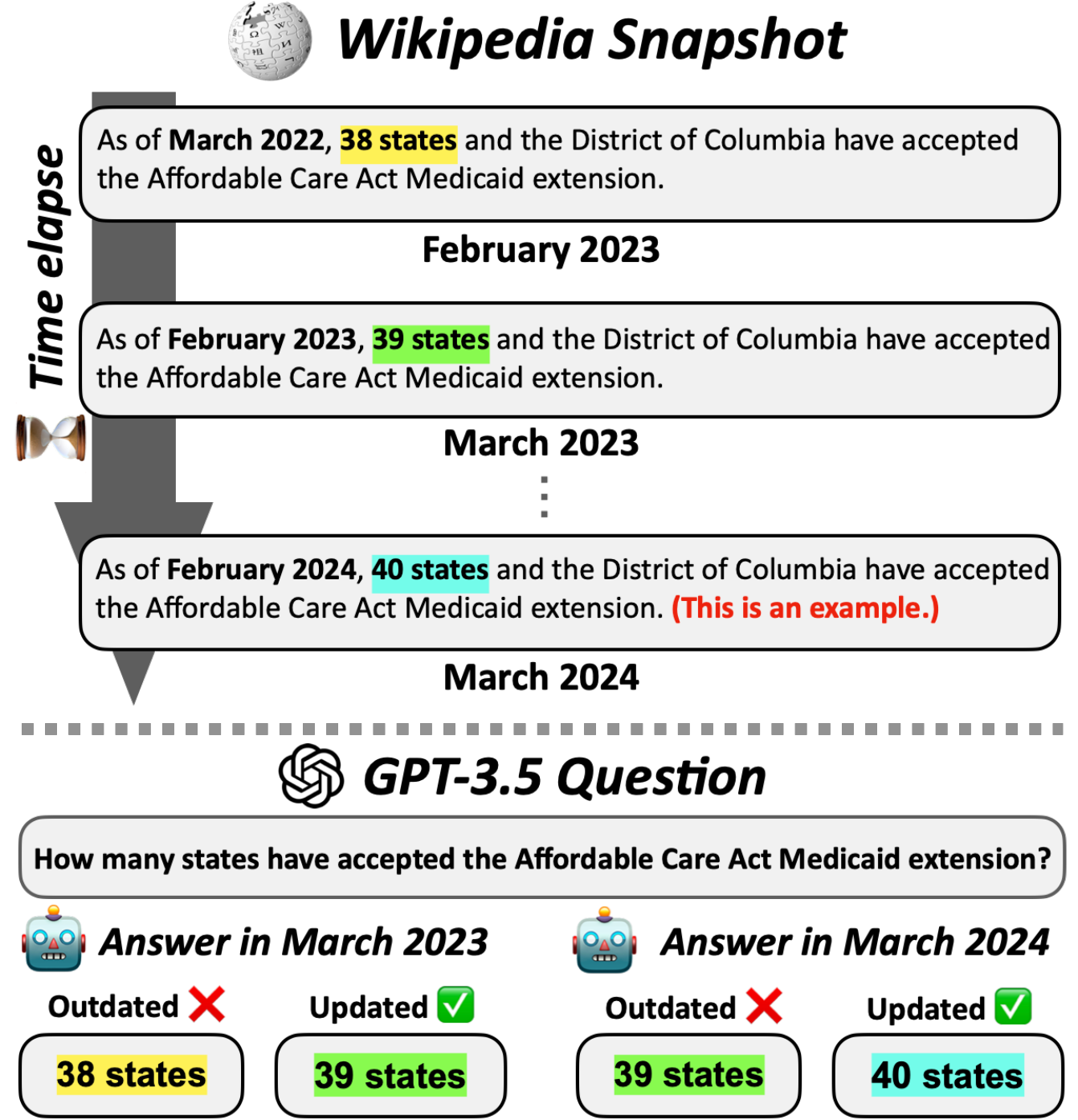}
    \caption{An overview of our evaluation benchmark, EvolvingQA. Our benchmark employs LLM to generate question-answer pairs based on the changes in Wikipedia's snapshots, effectively capturing the temporal evolution of the knowledge base.}
    \label{fig:main}
\end{figure} 

Meanwhile, previous research has shown that language models pre-trained on reliable knowledge sources such as Wikipedia can substitute knowledge bases by storing knowledge in their parameters and be applied to various downstream tasks~\citep{petroni2019language, roberts2020much}. 
\begin{table*}[t!]
    \centering
    \resizebox{0.8\linewidth}{!}{
    \begin{tabular}{c|ccccc}
    \toprule
     \multirow{2}{*}{\textsc{Attribute}} & \textbf{EvolvingQA} & CKL & TemporalWiki & StreamingQA & RealTimeQA \\ 
     & (Ours) & \citep{jang2021towards} & \citep{jang2022temporalwiki} & \citep{livska2022streamingqa} & \citep{kasai2022realtime} \\ \midrule
    \textsc{Edited knowledge} & \cmark & \cmark & \xmark & \xmark & \xmark \\
    \textsc{Automatic construction} & \cmark & \xmark & \cmark& \xmark & \xmark\\
    \textsc{\# of timestamps} & $6$ (Unlimited) & $2$ & $4$ (Unlimited) & 4 & (Unlimited)\\
    \textsc{Available Tasks} & QA & Slot-filling & Slot-filling & QA & QA \\
    \bottomrule
    \end{tabular}}
    \caption{Comparison of our benchmark and existing benchmarks for temporal alignment. Detailed descriptions for each attribute are presented in Appendix \ref{appx:benchmark_comparison}.}\label{tab:benchmark_comparison}
\end{table*}
To keep these models up to date with evolving world knowledge, it is desirable to apply continuous pre-training rather than periodically re-training from scratch.

Continued learning of existing models over sequential time-varying data remains one of the critical challenges in machine learning and has been widely discussed in previous literature, often referred to as continual learning (CL)~\citep{ThrunS1995,LiZ2016eccv,LeeS2017nips,wang2022learning} or lifelong learning.
This learning paradigm addresses the problem of learning on multiple tasks/data sequentially, assuming that the data from the previous session is inaccessible when starting the next training session. The primary goal is to preserve previously acquired knowledge while learning new concepts. 

%
However, in real-world scenarios, consistent accumulation of world knowledge while \textit{forgetting outdated knowledge is desirable} due to changes in world knowledge.
Models are required not only to learn \textit{new} information but also to forget or \textit{update} outdated information\footnote{\textit{New} knowledge refers to added knowledge which was previously nonexistent, while \textit{updated} knowledge refers to added knowledge which invalidates previous knowledge.}.
For example, the knowledge from 2017 that ``Donald Trump is the president of the US.'' became outdated in 2021, when it was substituted by the updated knowledge ``Joe Biden is the president of the US.''
Several benchmarks have been introduced to evaluate language models on temporally changing knowledge \citep{jang2021towards, jang2022temporalwiki, neelam2022benchmark, livska2022streamingqa, kasai2022realtime}. However, these fall short in providing holistic evaluations of knowledge preservation \textit{and} modification in the context of real-world applications.
While \citet{jang2021towards, jang2022temporalwiki} address both changed and unchanged knowledge, models are evaluated using template-based knowledge probing (i.e., LAMA task~\citep{petroni2019language}), which may not represent applicability in the real world.
\citet{kasai2022realtime} focus on evaluating new and updated knowledge, thereby failing to assess catastrophic forgetting of previous knowledge after updated knowledge acquisition.
We provide a comprehensive comparison of benchmarks in Table \ref{tab:benchmark_comparison}.

Our goal is to create a benchmark for holistic evaluation of the temporal adaptation capabilities of language models.
We propose \textbf{EvolvingQA}, a novel benchmark for pre-training and evaluating LMs over evolving Wikipedia data.
We propose an automated pipeline to construct our benchmark using LLMs, which allows us to extend our benchmark to many time steps and to easily update the benchmark into the future, as depicted in Figure \ref{fig:main}.
We use the question-answering (QA) task for downstream evaluation to measure continual learning that translates into real-world applicability.
We find that continual pre-training baselines (1) suffer from catastrophic forgetting and (2) fail to forget outdated knowledge (3) or incorporate updated knowledge, highlighting the relevance of our benchmark.
We provide comprehensive analyses on why and how such circumstances occur.

Our contributions are as follows:
\begin{itemize}
    \item We propose a new benchmark to evaluate LMs on preserving time-invariant knowledge while integrating changes through continual pre-training. Our benchmark incorporates open-domain question-answering, which is an intuitive and practical downstream task. Our dataset construction pipeline is automated by using LLMs, which can be generated at low cost.
    \item Our experimental results on EvolvingQA show that the baselines struggle to learn updated knowledge and forget previously learned outdated knowledge. 
    \item We provide in-depth analyses on why and how the existing baselines fail to predict updated information. The language models especially struggle to update numerical or temporal knowledge, because the models' gradient is not significant enough to forget outdated knowledge when learning updated knowledge.
\end{itemize}

%% file: tex/04_dataset.tex
\begin{figure*}[t]
    \centering
    \includegraphics[width=\linewidth]{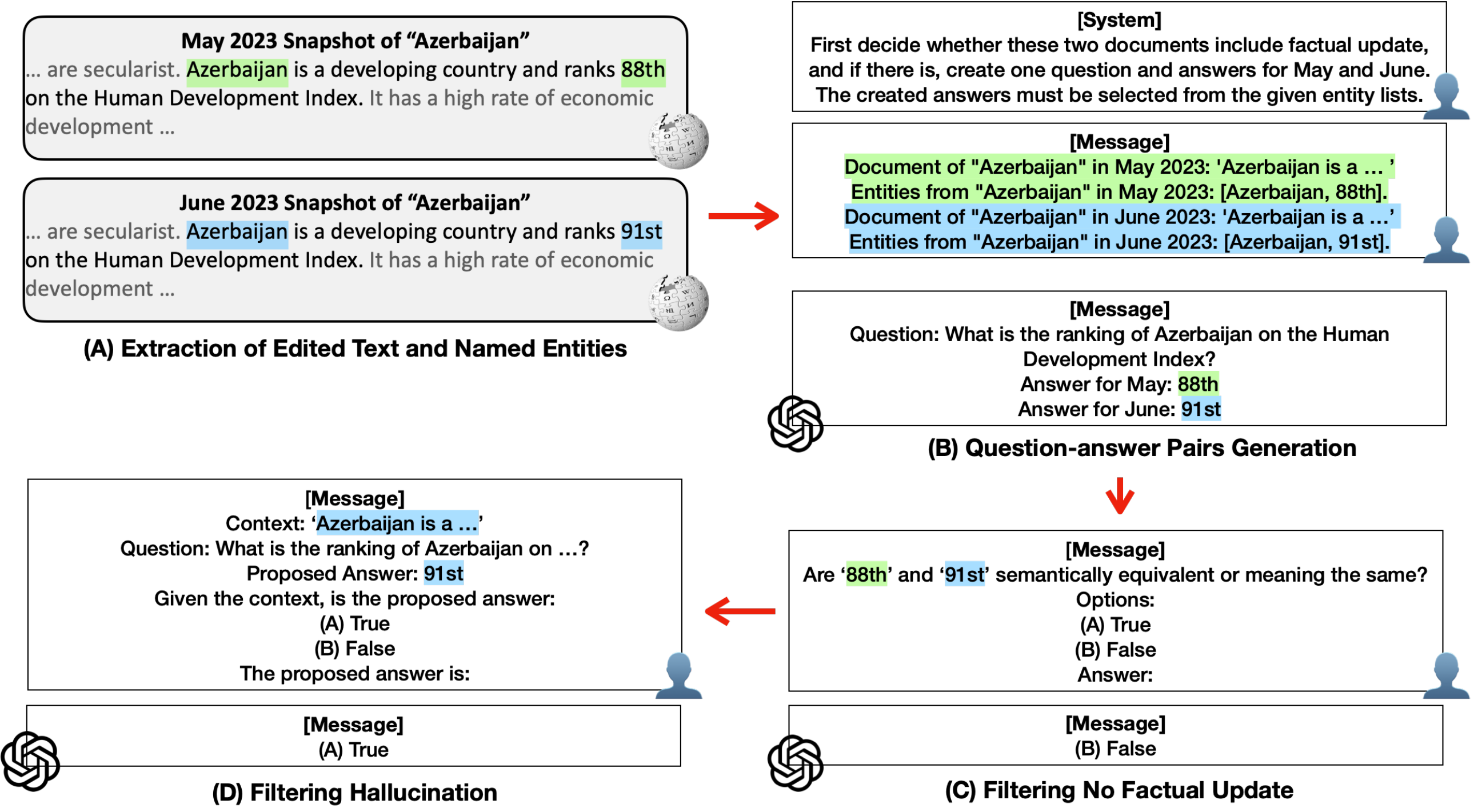}
    \caption{Construction pipeline of \edited. The final question-answers pair after filtering processes in this Figure is included in \textsc{Edited06}. The full description of the pipeline is in Section \ref{subsec:benchmark} and Appendix \ref{appx:eval_detail}.}
    \label{fig:edited_pipeline}
\end{figure*}
\section{EvolvingQA}
In this section, we introduce EvolvingQA, a novel benchmark for evaluating LM's ability to forget and update dynamically evolving knowledge. 
EvolvingQA is divided into continual pre-training corpora and evaluation dataset. For continual pre-training corpora, we collect consecutive Wikipedia snapshots and conduct heuristic filtering. For evaluation dataset, we collect a QA dataset through automatic generation and validation using LLM. Since both pre-training and evaluation data can be collected automatically, EvolvingQA could be extended to future time steps. 
\subsection{Continual Pre-training Dataset}
We collect \textsc{Changed} sets, pre-training corpora consisting of changes between two consecutive Wikipedia snapshots.
We exclude Wikipedia articles with minimal updates from \textsc{Changed} sets, focusing more on knowledge that have undergone sufficient changes.
Specifically, we only select Wikipedia articles that the updated part is more than the length of 500 characters as our continual pre-training dataset. We call these resulting subsets a \textsc{Changed} set.
For example, the \textsc{Changed03} set includes parts of Wikipedia articles of March 2023 that were modified from February 2023.
The number of topics in different corpus from each time step is shown in Table \ref{tab:stats_topic}. 
 We process \textsc{Changed} to follow T5 pre-training objective. Particularly, following \citet{roberts2020much}, we use salient span masking which set preparing the input as a text in which named entities and dates~\footnote{We use \texttt{en\_core\_web\_trf model} to extract named entities and dates provided from spaCy (\url{https://spacy.io/}).} are masked, and the output is then set as the corresponding unmasked entities and dates.
A sample of input and output from \textsc{Changed03} is reported in Figure \ref{fig:changed_sample}. 

\subsection{EvolvingQA benchmark}\label{subsec:benchmark}
We construct a question-answering benchmark to measure the model's capability of answering correctly while learning temporally changing knowledge. 
To measure how the language models 1) prevent catastrophic forgetting of old knowledge, 2) acquire new knowledge, and 3) edit their outdated knowledge into updated knowledge, we construct \unch , \new , and \edited evaluation sets, respectively.

We extract parts of Wikipedia articles that are unmodified, new, and edited, using the \texttt{difflib} library. We then prompt GPT-3.5\footnote{We use GPT-3.5-turbo-0613 provided by OpenAI API.} to generate question-answer pairs using the extracted parts. GPT-3.5 is conditioned to select answers from the given named entities, to ensure short-form answers. Note that the named entities that we provide GPT-3.5 are the ones that our language model is learned to reconstruct during pre-training \textsc{Changed} sets. The generated question-answer pairs are provided to GPT-3.5 as input for further filtering. 

\begin{table}[t]
    \centering
    \resizebox{\linewidth}{!}{
    \begin{tabular}{cccccccc}
    \toprule
     Dataset & 03 & 04 & 05 & 06 & 07 & 08\\ \midrule
     \unch & $49,504$ & $49,504$ & $49,504$ & $49,504$ & $49,504$ & $49,504$ \\ 
    \new & $29,680$ & $32,954$ & $31,487$ & $32,845$ & $38,584$ & $32,559$\\
    \edited & $7,293$ & $2,259$ & $1,889$ & $1,708$ & $1,672$ & $8,462$\\ 
     \bottomrule
    \end{tabular}
    }
    \caption{The number of question-answer pairs for evaluation.}\label{tab:stats_eval}
\end{table}
\vspace{-10pt}

\paragraph{\unch} The \unch evaluation set aims to measure how well the models maintain the knowledge obtained initially, even after learning the series of upcoming knowledge. We gather Wikipedia articles from the February 2023 snapshot that have not altered during the next six months. We then utilize the unaltered parts to prompt GPT-3.5 as context to create question-answer pairs. We condition GPT-3.5 to select the ground truth answer to be one of the given entities that were masked for pre-training input. The resulting \unch set is used to evaluate models on all time steps.

In order to make language models answer a given question in a desired format, fine-tuning pre-trained models on question-answering task is required. We extract 80K additional question-answer pairs from unchanged topics to construct a fine-tuning dataset, and we ensure it is disjoint with \unch set.  Consequently, continually pre-trained models are fine-tuned using the 80K unchanged pairs, and then evaluated with \unch, \new, and \edited of the corresponding time step. The resulting statistic of our benchmark is reported in Table \ref{tab:stats_eval}.

\begin{figure}[!t]
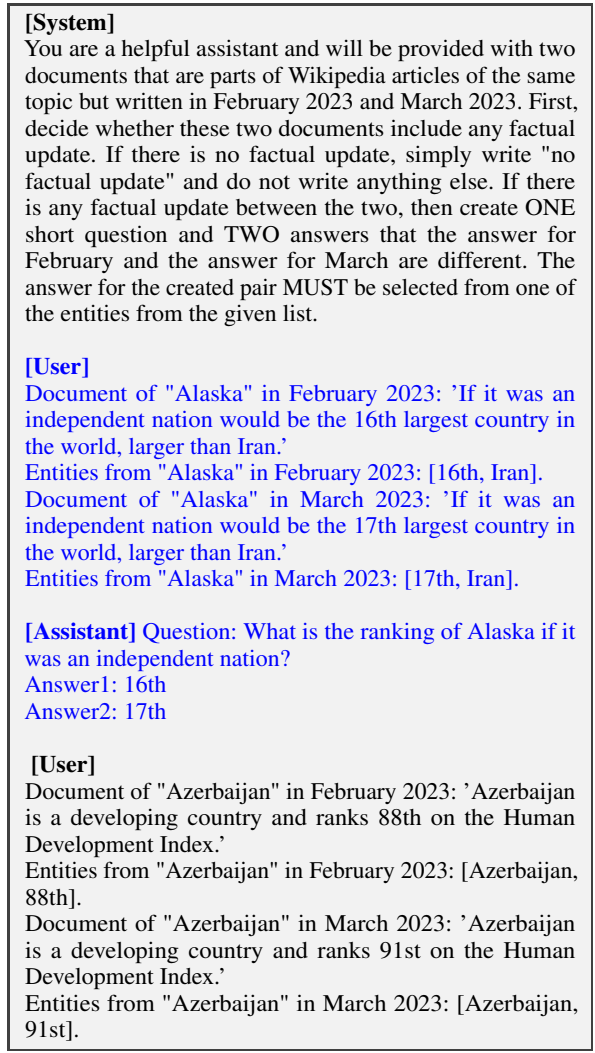

    \centering
    \begin{footnotesize}
    \begin{tcolorbox}[width=\linewidth, sharp corners=all, colback=gray!10, boxrule=0.1mm, size=small]
    \textbf{[System]}\newline
    You are a helpful assistant and will be provided with two documents that are parts of Wikipedia articles of the same topic but written in February 2023 and March 2023. First, decide whether these two documents include any factual update. If there is no factual update, simply write "no factual update" and do not write anything else. If there is any factual update between the two, then create ONE short question and TWO answers that the answer for February and the answer for March are different. The answer for the created pair MUST be selected from one of the entities from the given list.\newline\newline
    \textcolor{blue}{
    \textbf{[User]}
    \newline Document of "Alaska" in February 2023: 'If it was an independent nation would be the 16th largest country in the world, larger than Iran.'\newline
    Entities from "Alaska" in February 2023: [16th, Iran].\newline
    Document of "Alaska" in March 2023: 'If it was an independent nation would be the 17th largest country in the world, larger than Iran.' \newline
    Entities from "Alaska" in March 2023: [17th, Iran]. \newline\newline
    \textbf{[Assistant]}
    Question: What is the ranking of Alaska if it was an independent nation?\newline
    Answer1: 16th\newline
    Answer2: 17th\newline\newline}
    \textbf{[User]}
    \newline Document of "Azerbaijan" in February 2023: 'Azerbaijan is a developing country and ranks 88th on the Human Development Index.'\newline
    Entities from "Azerbaijan" in February 2023: [Azerbaijan, 88th].\newline
    Document of "Azerbaijan" in March 2023: 'Azerbaijan is a developing country and ranks 91st on the Human Development Index.' \newline
    Entities from "Azerbaijan" in March 2023: [Azerbaijan, 91st]. 
    \end{tcolorbox}
    \end{footnotesize}
    \caption{An example of the prompt we use in generating QA pairs in \edited set. The blue-colored messages are one-shot demonstration to make sure GPT-3.5 follow the instruction more accurately and generate question-answer instances in a desired format.}
    \label{fig:prompt}
    \vspace{-5pt}
\end{figure}

\paragraph{\new} The \new evaluation set shows how well the language models learn new knowledge that does not contradict the previously learned knowledge. We use \textsc{Changed} set of corresponding time steps to construct \new evaluation set. For example, to evaluate a model continually pre-trained until May 2023 (i.e., a model continually pre-trained from initial time step to \textsc{Changed05}), we use the \new05 set which consists of question-answer pairs extracted from \textsc{Changed05}. Similar to \unch, we prompt GPT-3.5 to create question-answer pairs, while conditioning answers should be selected from the given entities. 

\paragraph{\edited} The \edited evaluation set measures how the models forget outdated knowledge and learn updated knowledge when the previously learned knowledge gets outdated by the articles edited. The overview of our \edited construction pipeline is depicted in Figure \ref{fig:edited_pipeline}. In order to create question-answer pairs that reflect the edit of knowledge, we collect the revised parts of Wikipedia articles, and provide GPT-3.5 the original part (i.e., outdated part as of current time step) and the corresponding revised part (i.e., updated part as of current time step). The prompt we used to generate the QA pairs is described in Figure \ref{fig:prompt}. To filter out cases where the update only includes stylistic change or grammatical correction, we use system command to condition GPT-3.5 to determine if the context from two consecutive time steps does include factual updates. We also condition the answers should be one of the provided candidate entities for short and precise answers. Lastly, we provide GPT-3.5 one-shot example of question-answer generation for better alignment. The resulting \textsc{Edited} QA instance generated by GPT-3.5 includes a question, an \textsc{outdated} answer, and an \textsc{updated} answer.

After extracting question-answer pairs, we go through further filtering process to remove the hallucination and bias of GPT-3.5 by asking whether the answer is correct given the context and question, following \citet{kadavath2022language}. The details of prompts and filtering methods used in EvolvingQA construction pipeline are described in the Appendix \ref{appx:eval_detail}. 

%% file: tex/05_experiment.tex
\section{Experiment}

\begin{table*}[h!]
    \centering
    \resizebox{\linewidth}{!}{
    \begin{tabular}{cc|cccccc|ccccccc}
    \toprule
    \multirow{2.5}{*}{Method} & \multirow{2.5}{*}{Dataset} & \multicolumn{6}{c|}{EM} & \multicolumn{6}{c}{F1}\\ 
    \cmidrule(l{2pt}r{2pt}){3-8} \cmidrule(l{2pt}r{2pt}){9-14}
     & & 03 & 04 & 05 & 06 & 07 & 08 & 03 & 04 & 05 & 06 & 07 & 08\\
    \midrule
    \multirow{4}{*}{\textsc{Initial}} & \unch & 5.17 & 5.17 & 5.17 & 5.17 & 5.17 & 5.17 & 10.37 & 10.37 &  10.37 & 10.37 & 10.37 & 10.37\\
    & \new & 4.82 & 4.97 & 4.41 & 5.18 & 5.23 & 4.03 & 8.64 & 8.82 & 7.90 & 8.77 & 9.02 & 8.05 \\
    & \out$\downarrow$ & 2.30 & 2.19 & 2.68 & 2.21 & 2.80 & 2.65 & 7.30 & 7.15 & 7.88 & 6.99 & 7.71 & 7.58 \\
    & \up$\uparrow$ & 2.41 & 2.27 & 2.59 & 2.57 & 2.34 & 2.33 & 7.35 & 6.91 & 7.48 & 7.28 & 6.71 & 7.28\\
    \midrule
    \multirow{4}{*}{\textsc{Full}} & \unch & 3.78 & 3.62 & 3.37 & 3.33 & 3.28 & 3.11 & 8.41 & 8.20 & 7.95 & 7.86 & 7.79 & 7.66\\
    & \new & 5.23 & 4.64 & 4.27 & 4.78 & 4.68 & 3.43 & 9.45 & 8.69 & 8.22 & 8.56 & 8.44 & 7.53 \\
    & \out$\downarrow$ & 2.43 & 2.15 & 2.82 & 1.96 & 2.70 & 2.10 & 7.22 & 7.06 & 8.09 & 6.62 & 7.37 & 7.03\\
    & \up$\uparrow$ & 2.23 & 2.49 & 2.33 & 2.47 & 2.19 & 2.05 & 7.73 & 7.78 & 8.04 & 7.59 & 7.36 & 7.59\\
    \midrule
     & \unch & 4.64 & 4.55 & 4.44 & 4.40 & 4.43 & 4.45 & 9.47 & 9.44 & 9.40 & 9.37 & 9.35 & 9.40\\
    K-Adapter & \new & 5.52 & 5.42 & 4.83 & 5.29 & 5.42 & 4.25 & 9.83 & 9.64 & 8.80 & 9.41 & 9.59 & 8.83  \\
    \citep{wang2020k} & \out$\downarrow$ & 2.44 & 2.68 & 2.64 & 2.42 & 2.80 & 2.58 & 7.62 & 7.78 & 7.98 & 7.78 & 7.72 & 7.80\\
    & \up$\uparrow$ & 2.43 & 2.79 & 2.95 & 2.97 & 2.60 & 2.70 & 8.02 & 8.32 & 8.84 & 8.36 & 7.72 & 8.31 \\
    \midrule
     & \unch & 4.65 & 4.43 & 4.41 & 4.39 & 4.35 & 4.37 & 9.45 & 9.25 & 9.46 & 9.27 & 9.33 & 9.33 \\
    LoRA & \new & 5.57 & 5.32 & 4.93 & 5.31 & 5.46 & 4.13 & 9.75 & 9.51 & 9.06 & 9.34 & 9.71 & 8.56 \\
    \citep{hu2021lora} & \out$\downarrow$ & 2.64 & 2.53 & 3.04 & 2.77 & 2.65 & 2.55 & 7.80 & 7.42 & 8.40 & 7.96 & 7.88 & 7.87 \\
    & \up$\uparrow$ & 2.64 & 2.87 & 2.95 & 2.82 & 2.70 & 2.54 & 8.31 & 8.16 & 8.31 & 8.40 & 8.11 & 8.42 \\
     \midrule
     & \unch & 40.58 & 40.07 & 41.62 & 40.12 & 39.98 & 40.00 & 43.32 & 42.52 & 42.95 & 42.44 & 41.28 & 42.52 \\
    DPR & \new & 18.54 & 24.67 & 22.00 & 21.33 & 22.67 & 23.33 & 22.91 & 29.42 & 25.71 & 25.18 & 28.08 & 27.38\\
    \citep{karpukhin2020dense}& \out$\downarrow$ & 4.23 & 4.01 & 3.67 & 4.00 & 5.33 & 4.28 & 10.84 & 10.73 & 10.73 & 10.55  & 12.56 & 10.16\\
    & \up$\uparrow$ & 23.87 & 29.33 & 19.33 & 16.67 & 19.67 & 21.33 & 29.74 & 35.98 & 21.40 & 20.60 & 25.02 & 25.93\\
     \bottomrule
    \end{tabular}
    }
    \caption{The results of question answering task according to baseline methods. Exact match (EM) and F1 score are measured. Note that in ideal setting, the performance of \out should be as close to zero as possible if the model successfully forgets outdated knowledge. The result is from a single run.}\label{tab:main_table}
\end{table*}

\vspace{-3pt}
\subsection{Training Details}
\vspace{-1pt}
We utilize 737M T5-large~\citep{raffel2020exploring}, specifically \texttt{google/t5-large-ssm} pre-trained checkpoint from \citet{roberts2020much}. 
In EvolvingQA continual learning framework, we begin with an initial checkpoint (\textsc{Initial}), which is further pre-trained on the entire Wikipedia snapshot of February 2023. 
We then continual pre-train sequentially on \textsc{Changed} sets, using the learning rate of 1e-3 and gradient accumulation by 3 with a batch size of 5. 
To evaluate model's knowledge on each time step, we fine-tune continually pre-trained models on the QA train set composed of unchanged knowledge. We use 1e-5 for the learning rate with a batch size of 32 and train for a single epoch to avoid memorization. Then, we evaluate it on \textsc{Unchanged}, \textsc{New}, and \textsc{Edited} evaluation sets of each corresponding time step. During inference, greedy decoding is used, and we pre-process the decoded output and ground truth answer by changing it into lowercase and removing punctuation. This process is applied identically across all time steps.
\subsection{Baselines}
\paragraph{\textsc{Initial}} \textsc{Initial} is a starting checkpoint, before any continual pre-training on the following \textsc{Changed} sets. We pre-train T5-large using the entire Wikipedia snapshot of February 2023. 
The checkpoint of \textsc{Initial} serves as the initial checkpoint of all the other CL methods. 
\paragraph{\textsc{Full}} 
We start from \textsc{Initial} and continue pre-training on \textsc{Changed} sets sequentially. The full model is updated without freezing any parameter. This approach is similar to domain-adaptive pre-training proposed from \citet{gururangan2020don}.
\paragraph{K-Adapter}
K-Adapter \citep{wang2020k} is an architecture-based continual learning method, which trains additional adapters to the LM while freezing the original parameters. We use $k$=2 where the adapters are inserted after the second and the last layers. We only freeze the encoder part of encoder-decoder network and update decoder and adapters following \citet{jang2021towards}.
\paragraph{LoRA}
We implement parameter-efficient training method, LoRA \citep{hu2021lora}, which trains rank decomposition matrices of each layer while freezing the original parameter. We use $r$=4 and adapt $W_q$ and $W_v$ in self-attention layer. We only freeze the encoder part of encoder-decoder network and update decoder and LoRA module.
\paragraph{DPR}
We compare baselines with the retrieval-based method proposed by \citet{karpukhin2020dense}, which encodes passages into dense representations and retrieves context representations closest to the question representations. Namely, the most relevant context for each question is determined by calculating the dot product of the question embedding with all the context embeddings from the knowledge base. The retrieved contexts are used as context in open-book question answering. We fine-tune the pre-trained T5-large using QA pairs of unchanged knowledge (i.e., \textsc{Unchanged} pairs for fine-tuning) providing the context, to create the reader model. 
We use \texttt{facebook/dpr-ctx$\_$encoder-single-nq-base} and \texttt{facebook/dpr-question$\_$encoder-single} \texttt{-nq-base} models to create context and question embeddings, respectively.


\subsection{Results}\label{subsec:results}

\begin{figure*}[t!]
    \centering
    \begin{subfigure}[h]{0.45\linewidth}
    \includegraphics[width=\linewidth]{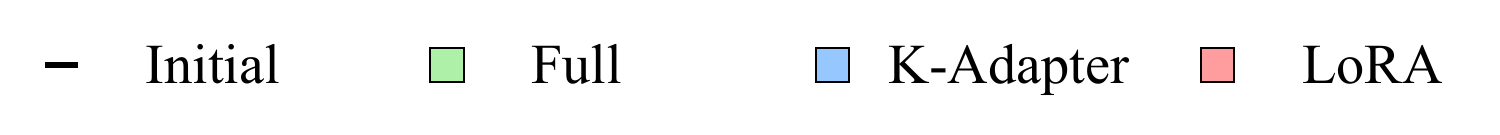}
    \end{subfigure}
    
    \begin{subfigure}[h]{0.25\linewidth}
    \includegraphics[width=\linewidth]{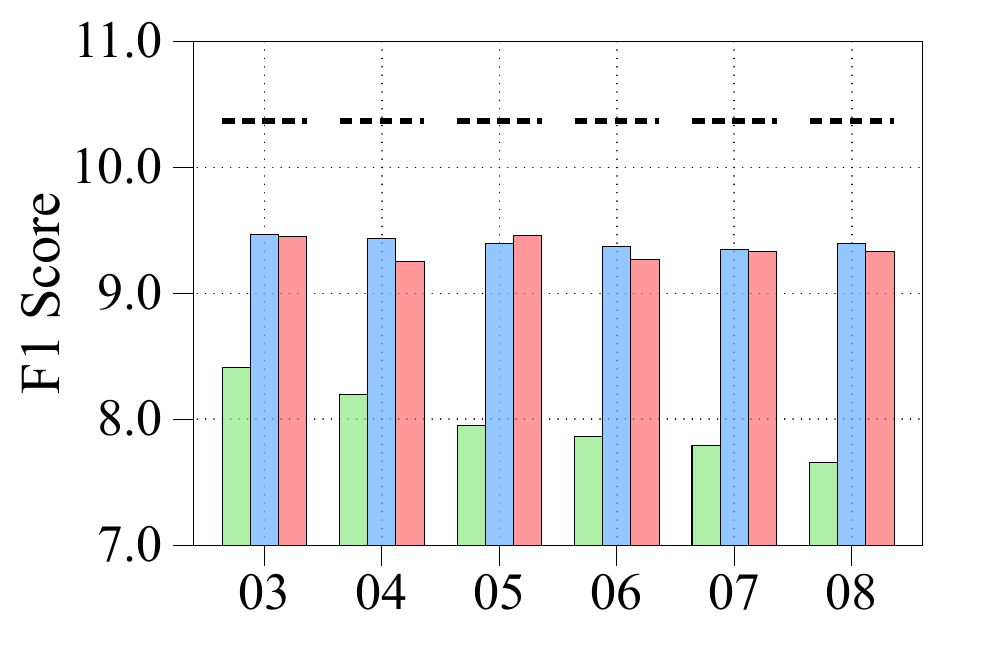}
    \caption{\textsc{Unchanged}}
    \end{subfigure}    
    \begin{subfigure}[h]{0.243\linewidth}
    \includegraphics[width=\linewidth]{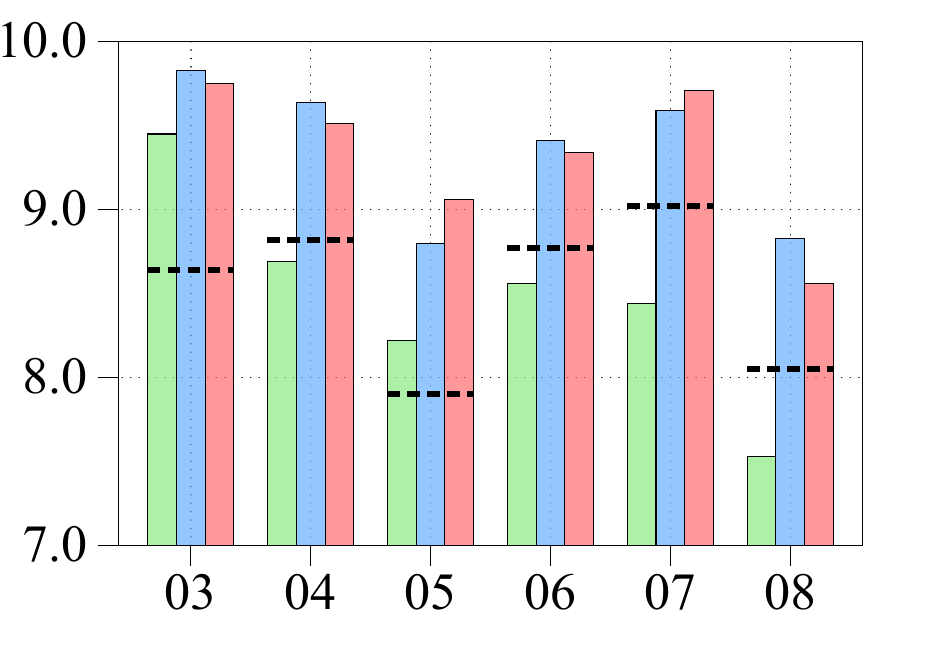}
    \caption{\textsc{New}}
    \end{subfigure}    
    \begin{subfigure}[h]{0.243\linewidth}
    \includegraphics[width=\linewidth]{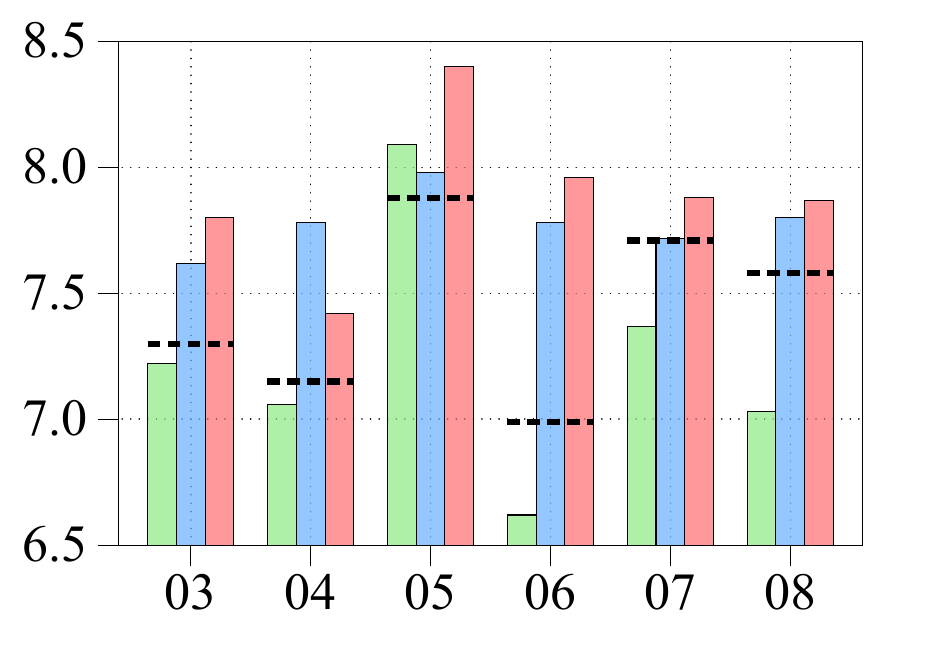}
    \caption{\textsc{Outdated} $\downarrow$}
    \end{subfigure}    
    \begin{subfigure}[h]{0.243\linewidth}
    \includegraphics[width=\linewidth]{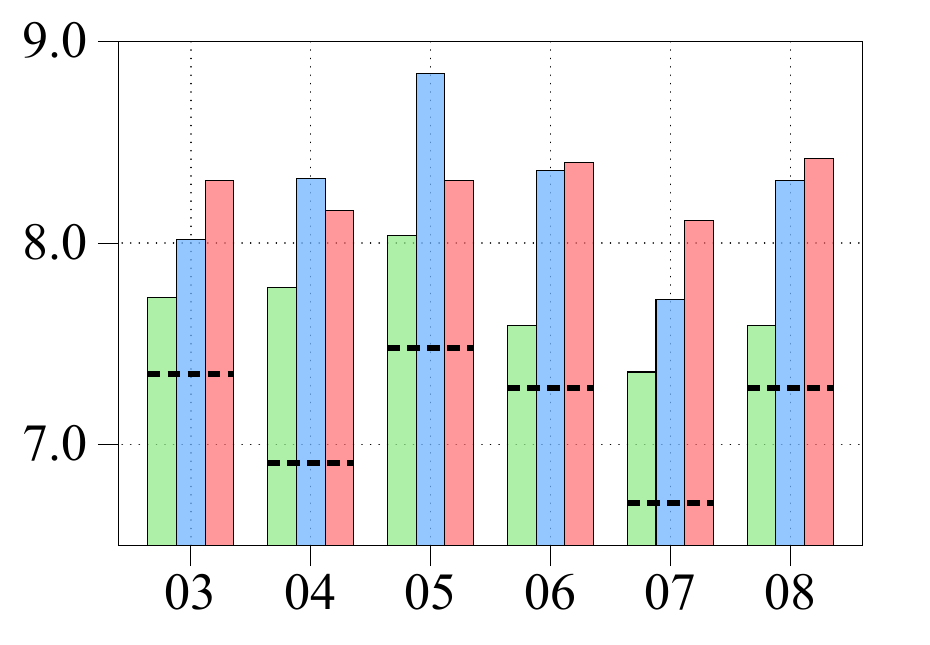}
        \caption{\textsc{Updated} $\uparrow$}
    \end{subfigure}    
    \caption{The bar plot that shows the trend of F1 scores through continual learning of \textsc{Changed} sets. Note that a single \unch set is used to evaluate on all time steps.}
    \label{fig:main_result}
\end{figure*}

Table \ref{tab:main_table} reports the result of baselines through sequentially learning Wikipedia articles from \textsc{Changed03} to \textsc{Changed08} starting from \textsc{Initial}. We measure Exact Match (EM) and F1 score, and F1 score is calculated by counting the common tokens between predicted answer and ground truth answer. 
We additionally provide visualization of the result of F1 scores in Figure \ref{fig:main_result}.
The result shows that all the CL baselines struggle with catastrophic forgetting, while \textsc{Full} forgets the unchanged knowledge the most.
\textsc{Full} also struggle from acquiring \new knowledge compared to other methods, and we conjecture that if the knowledge from different time steps is not learned with isolated parameters, it can result in blurring knowledge from different time steps. 
Meanwhile, K-Adapter and LoRA exhibit comparably high stability and plasticity, since they freeze the encoder and train with additional parameters.

In contrast, the overall performance in \out and \up presents that all baselines suffer from forgetting outdated knowledge and acquire updated knowledge. Ideally, the performance for \out should be close to zero when the model perfectly updated their knowledge. However, most of the baselines result in similar \out performance with \up performance. 

We also conduct additional experiment on shifting QA task into multiple choice answering, where \out and \up answer are two answer candidates. As shown in Table \ref{tab:result_select} in Appendix \ref{appx:multiple_choice}, the result also indicates that with more than 50\% of selecting \out answer, the models remain outdated. 

Meanwhile, DPR shows significant and meaningful result, where performance of \out is much lower than \up, thus demonstrating our benchmark's accuracy and faithfulness.


\subsection{Analysis on \edited Knowledge}
In this section, we delve into a thorough analysis of the reasons and mechanisms behind the failure of language models to update their information through the continual pre-training process.

\subsubsection{Gradients of \edited Knowledge}\label{subsec:edited_gradient}
\begin{figure}[t!]
    \centering
    \begin{subfigure}[h]{0.9\linewidth}
    \includegraphics[width=\linewidth]{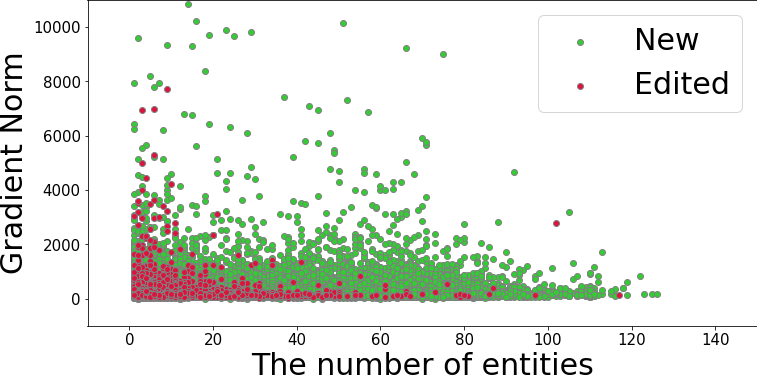}
    \end{subfigure}    
    \caption{The scatter plot of samples in \textsc{Changed03} according to the number of masked entities and gradient norm. Each dot indicates a sample from either \new knowledge or \edited knowledge in \ch. The $x$-axis shows the number of masked entities in a sample. The $y$-axis shows the Frobenius norm of weight gradients of each sample.}
    \label{fig:result_analysis2}
\end{figure}

We conduct an analysis to observe the differing trends in gradient updates when the model processes new or edited information during continual pre-training. Figure \ref{fig:result_analysis2} illustrates the Frobenius norm of the model's weight gradients when exposed to newly introduced or updated knowledge, specifically using inputs from the \textsc{Changed03} set. This is calculated based on the \textsc{Initial} checkpoint, encompassing gradients across all parameters. Similar trends are observable in other time steps as depicted in Appendix \ref{appx:result_analysis2_other}.
Notably, when the model is fed with updated knowledge (red color), the norms of the weight gradients are considerably smaller and closer to zero, in contrast to when it processes new knowledge (green color). This suggests that the model's gradient updates are less significant to forget the outdated information when trained with updated information. We hypothesize that this is because the updated information closely resembles the form of the previously learned information, rendering it more recognizable to the model.
\begin{figure*}[t!]
    \centering
    \includegraphics[width=0.4\linewidth]{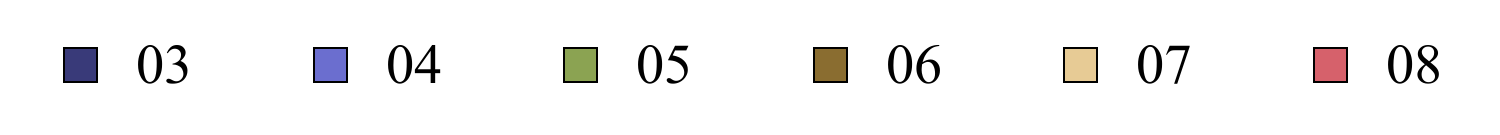}
    \vspace{2pt}
    
    \begin{subfigure}[h]{0.35\linewidth}
    \includegraphics[width=\linewidth]{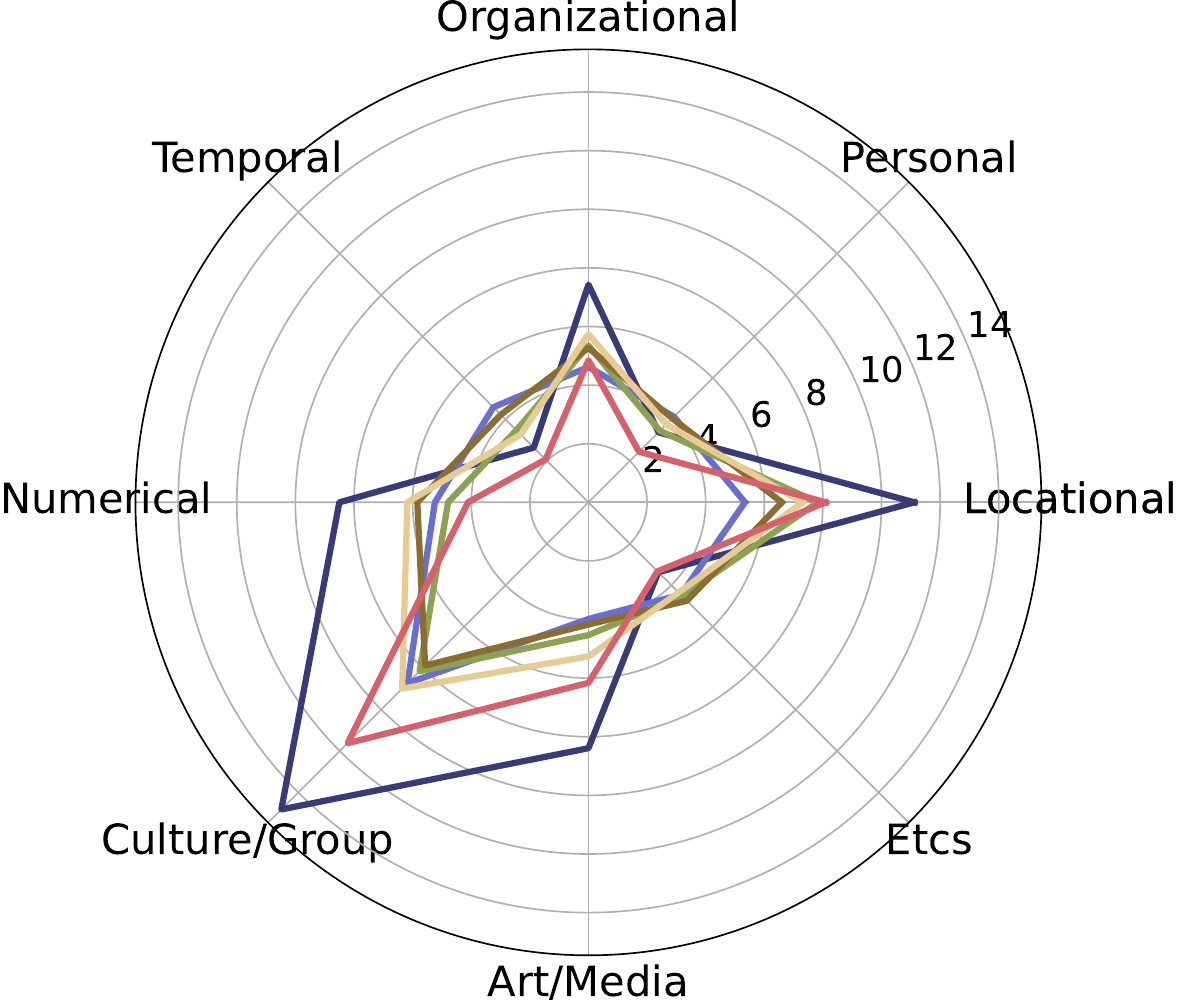}
        \caption{\new}
    \end{subfigure}    
    \hspace{10pt}
    \begin{subfigure}[h]{0.35\linewidth}
    \includegraphics[width=\linewidth]{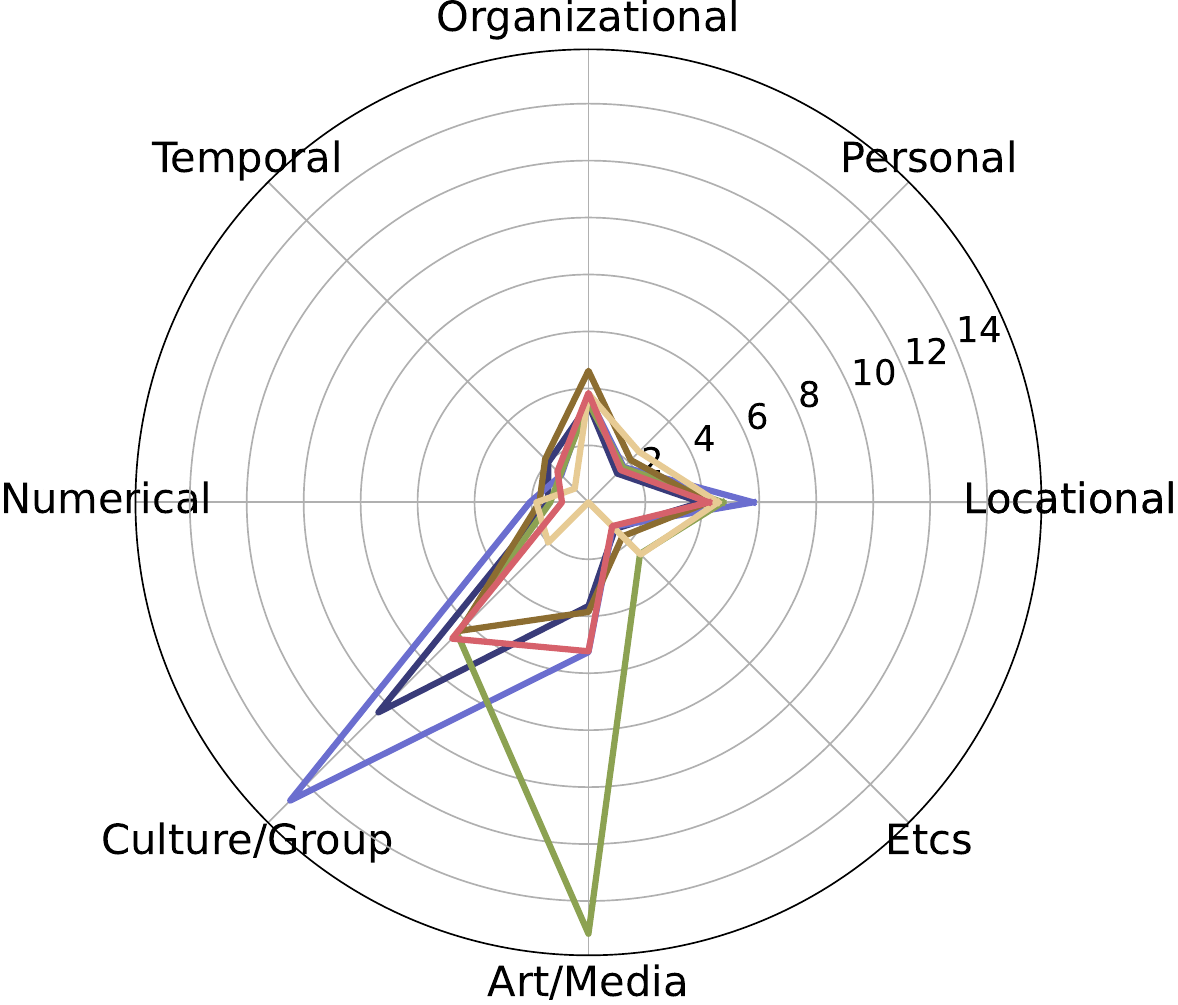}
        \caption{\edited}
    \end{subfigure}
    \caption{The analysis of EM score according to QA category. The result of each time step is shown in different colors.}
    \label{fig:result_analysis}
\end{figure*}
\subsubsection{Quantitative Analysis of \edited knowledge}
To investigate the types of knowledge that LMs struggle to update, we categorize QA instances into eight distinct types. Figure \ref{fig:result_analysis} presents the EM scores for each category, using the \textsc{Full} method, evaluated on \new and \edited sets.

As illustrated in Figure \ref{fig:result_analysis} (a), the distribution of EM scores across different categories in the \new set remains consistent across all time steps. Notably, models that have undergone continual pre-training show enhanced performance in the Culture/Group, Locational, and Art/Media categories. However, as depicted in Figure \ref{fig:result_analysis} (b), the models exhibit challenges in accurately predicting knowledge within the numerical and temporal categories for the \edited set, with EM nearing zero across all time steps. This suggests a notable deficiency in the models' ability to effectively update numerical or temporal information. The descriptions of each category are explained in Appendix \ref{appx:category_detail}.

%% file: tex/06_discussion.tex
\section{Discussion}
\subsection{Knowledge Change in Wikipedia}
Wikipedia, a widely-used online encyclopedia, exemplifies collective intelligence with its open editing system. Its monthly snapshots enable tracking of article changes, forming the basis of our dataset. These changes are categorized into three types: (1) updates with recent news or facts, (2) additions or corrections of existing information, and (3) grammatical corrections.
\paragraph{Updates with Recent News} involve adding current events or new discoveries, reflecting the evolving nature of world knowledge. It is the most crucial part that our benchmark aims to encompass. Note that such update does not always reflect real-time news immediately.
\paragraph{Additions or Corrections of Existing Information} are frequent, involving updates to historical events or figures. While not always reflecting current events, it is important to consider these modification. For the cases where models have learned erroneous or private data, ensuring models to remain accurate and respectful of privacy concerns is significant and challenging. Continuous information revision is key to the development and ethical integrity of language models.
\paragraph{Grammatical Corrections} are minimized in our EvolvingQA dataset using heuristic algorithms and LLM validation. However, a few instances of grammatical or spelling updates remain in our dataset. Advanced models like GPT-4 could further reduce such cases.

\subsection{When does \textsc{Initial} accurately answer \textsc{New} and \textsc{Updated} questions?}
\begin{figure}[t!]
    \centering
    
    \includegraphics[width=\linewidth]{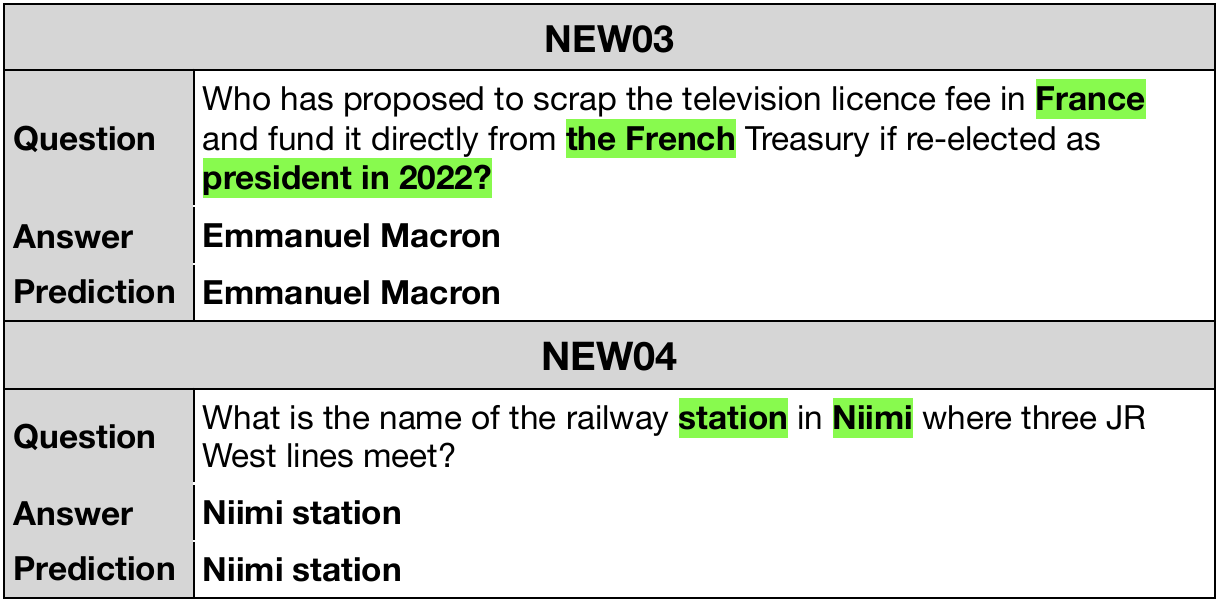}
    \caption{Samples that \textsc{initial} answers correctly from questions in \textsc{new}.}\label{tab:initial_correct}
    \label{fig:new_examples}
\end{figure}

Although \textsc{Initial} was trained on Wikipedia's February 2023 data, it can answer questions about newer or revised information. This may be due to several reasons. First, some questions contain the answers within them, allowing correct responses without updated knowledge. For example, as shown in a sample from \textsc{New04} in Figure \ref{fig:new_examples}, the answer is already in the question. Second, predictions can be made based on previous knowledge. For instance, in a sample from \textsc{New03}, certain answers might be inferred from keywords like "France" and "president". Lastly, the T5 pre-training, which includes various sources beyond Wikipedia, might have provided the model with relevant background knowledge.
The effectiveness of \textsc{Initial} in handling these questions also relates to the question difficulty level, our benchmark includes questions that are too easy to answer. This limitation can be improved with more advanced models like GPT-4 or by adjusting the question difficulty settings.

\subsection{Can Retrieval Replace Continual Learning?}
Section \ref{subsec:results} reveals that DPR outperforms CL baselines in performance. However, this doesn't necessarily undermine the value of continual learning on language model. In EvolvingQA, it is easy for retrieval methods to search for relevant context because the questions often repeat words from their context, leading to a high overlap. For instance, the question "How many states have accepted the Affordable Care Act Medicaid extension?" directly mirrors its context's phrasing, when the context is "...39 states have accepted the Affordable Care Act Medicaid extension...". 
Additionally, since the questions merely seek specific facts, the model simply reads and identifies the apparent answer in the context (i.e., executing one-hop prediction).

Continual learning remains crucial for language models, especially for real-world applications requiring complex reasoning and deep subject understanding. Language models need to integrate and apply their intrinsic knowledge to complex tasks, a capability beyond the scope of retrieval methods. Future research can focus on creating CL benchmarks that evaluate language models' ability to logically process and update knowledge.

\subsection{Is Closed-Book QA the Best Way to Assess the Knowledge of Model?}
In our research, we utilize the closed-book QA (CBQA) task to assess the knowledge of models. This method, however, requires careful consideration to determine its effectiveness in assessing a model's knowledge. For instance, there's a distinction between what a language model knows and how it responds, implying that CBQA results may not fully capture a model's inherent knowledge~\citep{lewis2020question, jiang2020can}.
Lastly, the current evaluation metrics, EM and F1, relies on lexical matching, and has limitations in verifying the accuracy of the model's predictions \citep{jiang2020can, risch2021semantic, bulian2022tomayto, kamalloo2023evaluating}. 
While our work is in early stages of research on continual learning for language models, we anticipate that considering such factors will enable the creation of benchmarks that are closer to optimal in the future. This direction is left as a promising avenue for future research. 

%% file: tex/03_related.tex
\section{Related Works}
\paragraph{Temporal Continual Learning Benchmarks in NLP}
\citet{zhang2021situatedqa} and \citet{kasai2022realtime} introduced QA datasets for temporal or geographical adaptation, but require manual annotation and disregard continual learning scenario. \citet{jang2022temporalwiki} constructed benchmark to reflect Wikipedia's dynamically changing knowledge in an automated manner, but they did not include an evaluation setting to measure updating outdated knowledge. \citet{jang2021towards} and \citet{livska2022streamingqa} proposed CL benchmarks relying on expert annotation and filtering, resulting in few timestamps and remaining static from the time it was created. 

\paragraph{Continual Learning and Model Editing}
There is an increasing interest in continual learning for language models, particularly focusing on domain-incremental and task-incremental learning~\citep{chen2020recall, qin2022elle, dhingra2022time,razdaibiedina2023progressive,chen2023lifelong,cole2023salient}. However, the area of temporally evolving CL and the term of forgetting outdated knowledge remains relatively under-explored. In the context of model editing~\citep{de2021editing, mitchell2021fast, meng2022locating, meng2022mass, huang2023transformer}, which is primarily aimed at updating and rectifying errors in language models, existing research often overlooks scenarios involving sequential updates. Moreover, the focus predominantly remains on updating extant knowledge, with less attention given to the acquisition of entirely new information.
More detailed related works are available in Appendix \ref{sec:additional_related}.


%% file: tex/07_conclusion.tex
\section{Conclusion}
Our research shed light on the importance for LMs capability of dynamically accumulating and revising information to reflect the continual evolution of world knowledge, which were under-explored in previous studies. Our proposed EvolvingQA benchmark includes evaluation for the adaptability of LLMs to such continual changes, revealing significant deficiencies in current models' abilities to forget and update outdated knowledge, especially in numerical and temporal data. Our findings show that this is due to the ineffectiveness of gradient update in managing updated knowledge. We hope that our work acts as a cornerstone for future research aiming to bridge the existing gaps in LLMs' temporal adaptation capabilities.

%% file: tex/09_limitation.tex

\paragraph{Limitation} Our study's limitations include the EvolvingQA dataset's lack of real-time updates. As Wikipedia updates monthly, there's a gap between current events and their reflection in the dataset. Additionally, using a single LLM for dataset construction and filtering processes introduce noise. LLMs can hallucinate and generate inaccurate data, and validation using the same LLM may not be possible to completely eliminate such risks. Though usage of advanced models such as GPT-4 and different validation model may mitigate this, it remains a concern. Furthermore, the overall performance is low, since closed-book QA itself is a very challenging task, and this can be alleviated by training models with larger capacities \citep{roberts2020much}. Finally, our framework do not allow control over question difficulty, affecting the evaluation results depending on the complexity of the questions. This might be addressed with refined prompting or additional pre-processing.

%% file: tex/08_ethics.tex
\paragraph{Ethics Statement} In the development and evaluation of our benchmark, we adhered to rigorous ethical standards concerning the use of data and the potential impacts of our research.  Our approach to continual pre-training and knowledge updating was designed to avoid the perpetuation of temporal biases, inaccuracies, or outdated information. We acknowledge that our benchmark and language models trained on it can be susceptible to reflecting societal biases present in training data. We will make every effort and take all possible measures to minimize and avoid such risks to the best of our ability.

%% file: tex/10_ack.tex
\paragraph{Acknowledgement} 
This work was supported by Institute of Information \& communications Technology Planning \& Evaluation (IITP) grant funded by the Korea government(MSIT) [No.2022-0-00641, XVoice: Multi-Modal Voice Meta Learning, 90\%] and [No. 2019-0-00075, Artificial Intelligence Graduate School Program (KAIST), 10\%]. 

%% file: tex/appx_01.tex
\newpage
\vspace{-5pt}
\section{Dataset Details}
\begin{table*}[h]
    \centering
    \resizebox{0.85\linewidth}{!}{
    \begin{tabular}{c|ccccccc|}
    \toprule     
     Time step& 
     \multirow{2}{*}{03}& 
     \multirow{2}{*}{04}& 
     \multirow{2}{*}{05}& 
     \multirow{2}{*}{06}& 
     \multirow{2}{*}{07}& 
     \multirow{2}{*}{08}\\ 
    (Month, 2023)& & & & & & \\ 
    \midrule
    Entire snapshot & $16,887,309$ & $16,918,791$ & $16,966,779$ & $16,997,214$ & $17,108,808$ & $17,233,540$\\ 
    \textsc{Changed} w/o filtering & $337,868$ & $353,934$ & $357,598$ & $362,606$ & $347,970$ & $361,699$\\
    \textsc{Changed} & $61,176$ & $65,780$ & $64,140$ & $66,938$ & $63,946$ & $68,075$\\ 
     \bottomrule
    \end{tabular}}
    \caption{The number of articles in \textsc{Changed} sets.}\label{tab:stats_topic}
\end{table*}

We collect Wikipedia snapshot from February 2023 to August 2023. For snapshot from February 2023, we use entire articles to pre-train \textsc{Initial}. We extract changes from two consecutive snapshots, and filter out articles that do not include much edited parts. The number of articles from each process is reported in Table \ref{tab:stats_topic}. 

\begin{figure*}[h!]
    \centering
    \includegraphics[width=\linewidth]{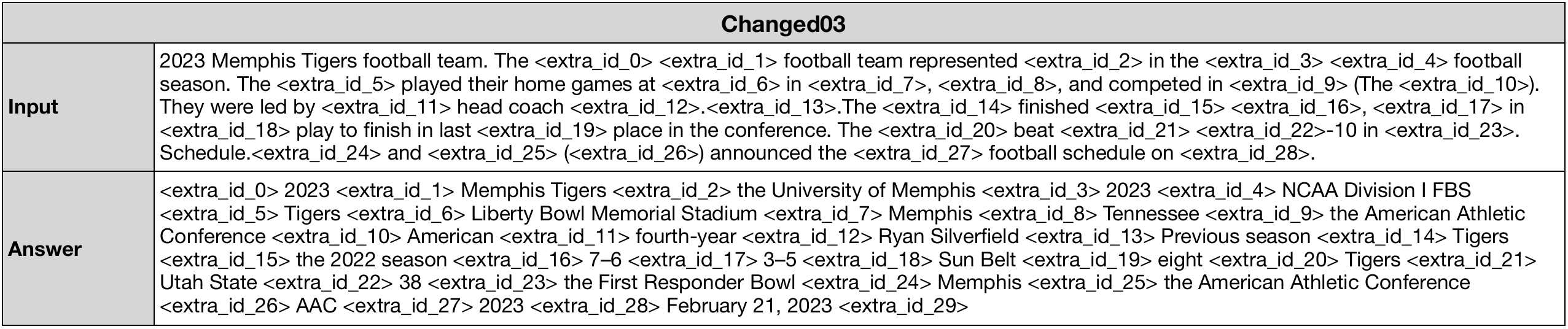}
    \caption{A sample of input and output from \textsc{Changed03}.}
    \label{fig:changed_sample}
\end{figure*}

To create \textsc{Changed} sets from these resulting articles, we use salient span masking, and a sample from \textsc{Changed03} set is shown in Figure \ref{fig:changed_sample}. In the input, named entities and dates are masked, and the output contains the masked entities. 

\section{Comparison of EvolvingQA with Other Benchmarks}\label{appx:benchmark_comparison}
Table \ref{tab:benchmark_comparison} reports the comparison between EvolvingQA and the existing benchmarks for temporal alignment. \textsc{Edited knowledge} denotes evaluation on updated and outdated knowledge, and \textsc{Automatic construction} denotes benchmark construction can be automated without human annotation. \textsc{\# of time steps} shows available time steps of the benchmark, while (Unlimited) denotes whether the construction framework can be applied dynamically to future time steps. \textsc{Available Tasks} shows benchmark's downstream task. 
Our benchmark have significant advantages including evaluation of edited knowledge, ability to be constructed automatically with unlimited number of time steps, and question answering as practical downstream task. 

\section{\edited Construction Details}\label{appx:eval_detail}
In this section, we provide details on construction of EvolvingQA, especially about constructing \edited set. Note that constructing \edited requires a lot of filtering and refinement process, since Wikipedia update includes grammar and error correction, so the update may not include factual update. Therefore, we go through multiple process of filtering and extensive prompt engineering to obtain QA pairs that actually reflect factual update. 

The prompt used to generate \edited is described in Figure \ref{fig:prompt}. Note that [System]\footnote{The system message is used to control the behavior of the AI model, such as by providing specific instructions.}, [Assistant], and [User] indicate "role" when providing messages to GPT-3.5 through API. Below are the examples of prompts we use in every step of construction pipeline when validating \edited set.

\subsection{Filtering No Factual Update}
The extracted QA instances still includes a number of instances that the outdated answer and the updated answer are written different, but actually the same. 
To filer out these cases, we prompt as below:

\begin{footnotesize}
\begin{tcolorbox}[width=\linewidth, sharp corners=all, colback=gray!10, boxrule=0.2mm]
{
Are '28' and 'Twenty-Eight' semantically equivalent or meaning the same?\newline
Options: \newline
(A) True\newline
(B) False\newline
Answer:}
\end{tcolorbox}
\end{footnotesize}
For above example, GPT-3.5 reponses as (A) True, then we filter out this instance from the dataset. 
There may be potential noise in our filtering process when using multiple-choice prompts~\citep{zheng2023large}. We incorporate varied seeds and altering the order of options.

\subsection{Filtering Hallucination}
For some instances, GPT-3.5 make up question even though there are no sufficient information in the context that supports the question and answer. In this regard, to filter out hallucinated instances, we use prompt following \citet{kadavath2022language}:

\begin{footnotesize}
\begin{tcolorbox}[width=\linewidth, sharp corners=all, colback=gray!10, boxrule=0.2mm]
{
"Context of '{Commuter rail}': Indonesia, the Metro Surabaya Commuter Line, Prambanan Express, KRL Commuterline Yogyakarta, Kedung Sepur, the Greater Bandung Commuter\newline
Question: Which commuter rail system was removed from the list in April 2023?\newline
Proposed Answer: the Greater Bandung Commuter\newline
Given the context, is the proposed answer:\newline
(A) True\newline
(B) False\newline
The proposed answer is:"}
\end{tcolorbox}
\end{footnotesize}
For the above case, GPT-3.5 responded (B) False, then we excluded this instance from the dataset.


%% file: tex/appx_02_additional_exp.tex
\section{Evaluation on \edited Knowledge in Multiple Choice Setting}\label{appx:multiple_choice}
\begin{table}[h]
    \centering
    \resizebox{\linewidth}{!}{
    \begin{tabular}{ccccccccccccccc}
    \toprule 
    {Method} & {Knowledge} & \multicolumn{1}{c}{03} & \multicolumn{1}{c}{04} & \multicolumn{1}{c}{05} & \multicolumn{1}{c}{06} & \multicolumn{1}{c}{07} & \multicolumn{1}{c}{08}\\ 
    \midrule
    \multirow{2}{*}{\textsc{Initial}} & \out & 53.33 & 53.04 & 52.37 & 53.1 & 54.49 & 53.52 \\
    & \up & 46.67 & 46.96 & 47.63 & 46.9 & 45.51 & 46.48\\
    \midrule
    \multirow{2}{*}{\textsc{Full}} & \out & 52.21 & 51.94 & 51.61 & 50.78 & 53.41 & 52.4\\
    & \up & 47.79 & 48.06 & 48.39 & 49.22 & 46.59 & 47.6\\
    \midrule
    \multirow{2}{*}{K-Adapter} & \out & 52.08 & 51.11 & 49.73 & 51.13 & 54.08 & 51.69\\
    & \up & 47.92 & 48.89 & 50.27 & 48.87 & 45.92 & 48.31\\
    \midrule
    \multirow{2}{*}{LoRA} & \out & 52.07 & 50.59 & 50.94 & 51.13 & 53.87 & 52.4\\
    & \up & 47.93 & 49.41 & 49.06 & 48.87 & 46.13 & 47.6 \\ 
     \bottomrule
    \end{tabular}
    }
    \caption{The results of multiple choice setting on \edited knowledge according to baseline methods. }\label{tab:result_select}
\end{table}
Following previous studies \citep{brown2020language, sanh2021multitask}, we evaluate the baselines on \edited knowledge using multiple choice setting (i.e., rank classification), which is selecting the label option (i.e., either outdated or updated) with higher log-likelihood. Namely, the model computes the log  probability of both updated and outdated ground-truth answer and uses the higher one as the predicted answer. The log proability is calculated by summing the negative log softmax logits of the model on the tokens in ground-truth answer. The result reported in Table \ref{tab:result_select} shows that all the baselines fail to capture updated knowledge, and tend to be skewed more to outdated knowledge.

\section{Prompting Time Information}\label{subsec:TP}
\begin{figure}[h]
    \centering
    \begin{subfigure}[h]{0.3\linewidth}
    \includegraphics[width=\linewidth]{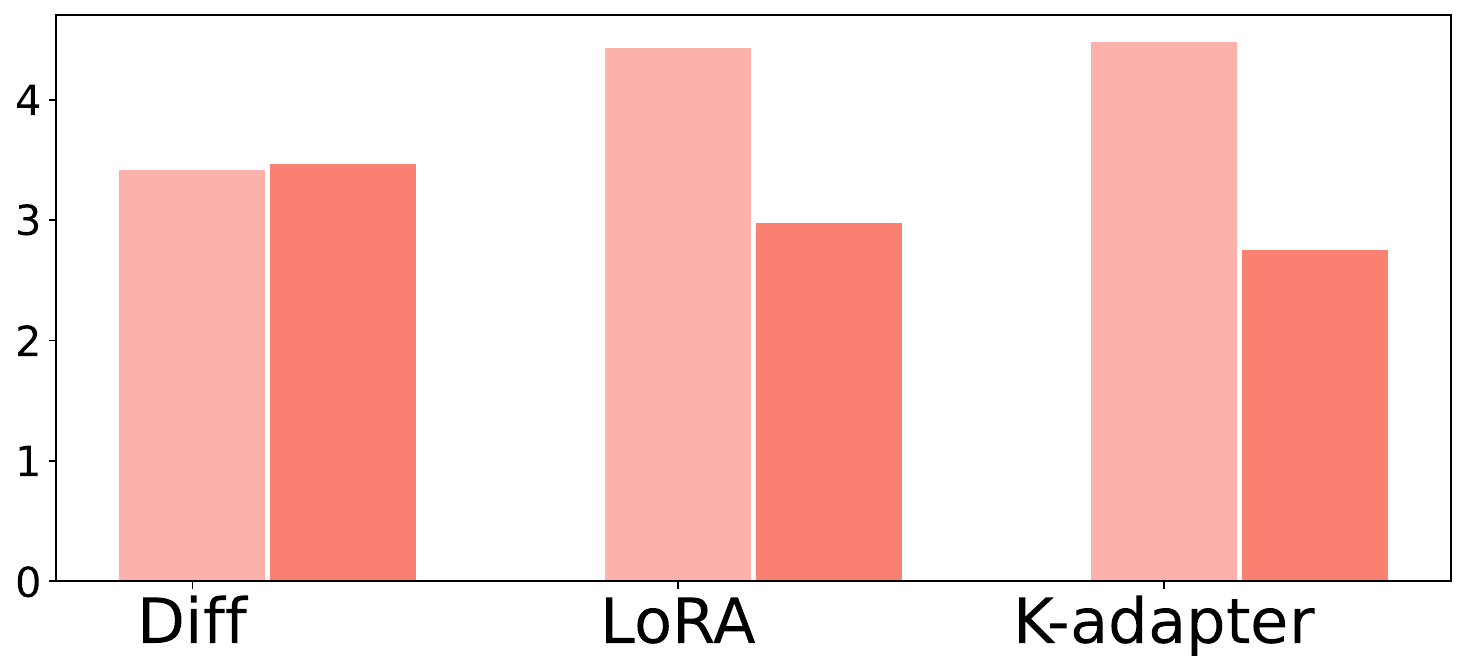}
        \caption{\unch}
    \end{subfigure}    
    \begin{subfigure}[h]{0.3\linewidth}
    \includegraphics[width=\linewidth]{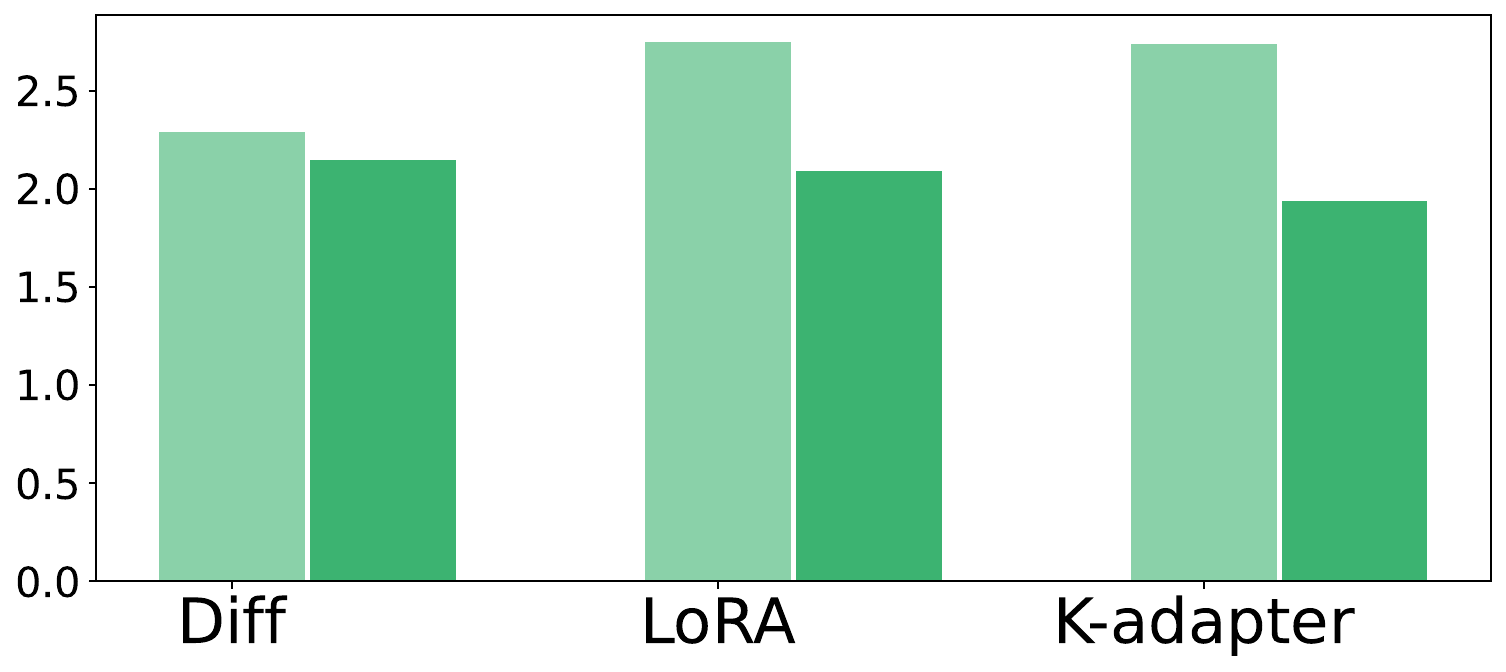}
        \caption{\edited}
    \end{subfigure}
    \begin{subfigure}[h]{0.3\linewidth}
    \includegraphics[width=\linewidth]{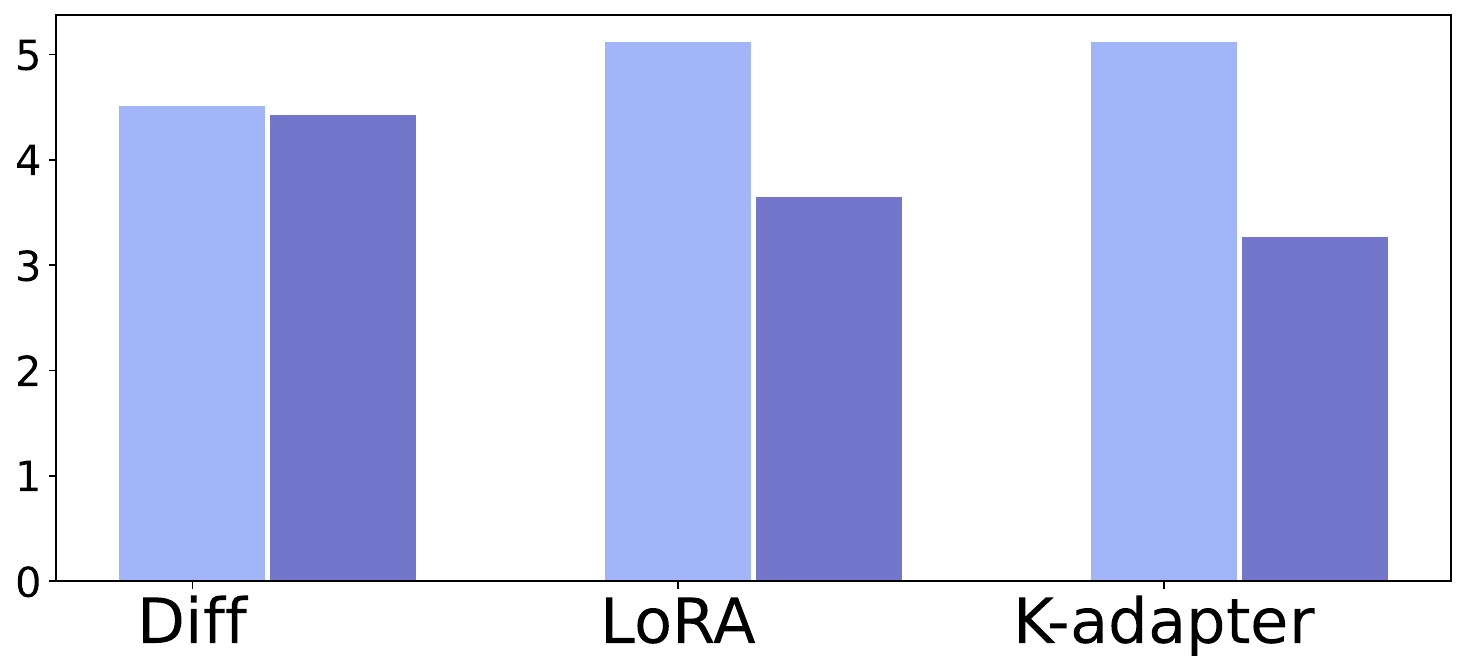}
        \caption{\new}
    \end{subfigure}
    \caption{Comparison between with and without adding time information into questions. The darker color indicates the result of adding time information. The EM score is averaged for all time steps. }
    \label{fig:tp_em}
\end{figure}

We add time information in the question, to see how e language model answers updated knowledge correctly after conditioning on time information. Specifically, when we test our models trained on \textsc{Changed05}, we then prepend "As of May 2023," to all the questions in \textsc{Unchanged05}, \textsc{New05}, and \textsc{Edited05}.
The result in Figure \ref{fig:tp_em} shows that inserting time information deteriorates the performance significantly. This is in line with \citet{kasai2022realtime} that in closed-book QA task, their date insertion method does not improve the performance. When we analyze the model's prediction when time information is given, the models tend to hallucinate more on temporal questions. Namely, when the models are asked to answer temporal questions asking dates, the models tend to reply with the date given as time information.

\section{Additional Results on Gradient Analysis of \textsc{Edited}}\label{appx:grad_result_other}
Figure \label{appx:result_analysis2_other} shows additional result of gradient norm analysis on \textsc{Changed04}. As in Section \ref{subsec:edited_gradient}, the result shows that gradient norm when learning edited knowledge is generally smaller than new knowledge. Note that we use instances from \textsc{Changed04} set using checkpoint from \textsc{Full} after pre-trained on \textsc{Changed03} and calculate gradients of the entire parameters. 
\begin{figure}[h!]
    \centering
    \begin{subfigure}[h]{\linewidth}
    \includegraphics[width=0.9\linewidth]{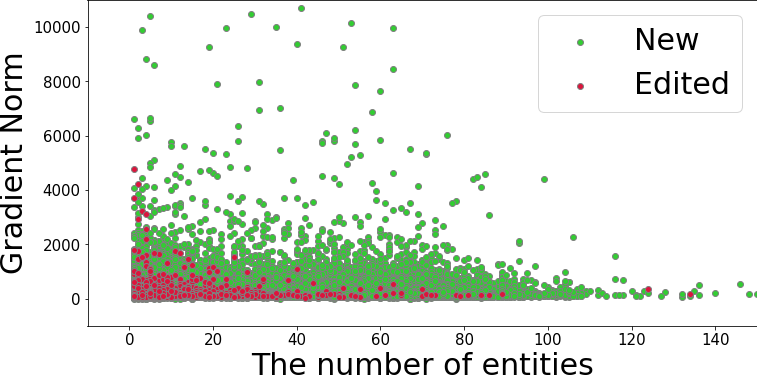}
    \caption{\textsc{Changed04}}
    \end{subfigure}    
    \caption{The scatter plot of samples from \textsc{Changed04} according to the number of masked entities and gradient norm. Each dot indicates a sample from either \new knowledge or \edited knowledge in \ch. The $x$-axis shows the Frobenius norm of weight gradients of each sample. The $y$-axis shows the number of masked entities in a sample.}
    \label{fig:result_analysis2_other}
\end{figure}

\section{Details about Categorization of QA samples}\label{appx:category_detail}
For categorization, we employ a Named Entity Recognition (NER) model to classify the categories of answers in our benchmark. The `Numerical' category encompasses answers identified as cardinal or ordinal numbers, quantities, and percentages. The `Temporal' category includes dates and times, while `Locational' encompasses geopolitical or geographical locations and facilities. `Organizational' refers to entities like organizations, and `Culture/Group' includes languages, laws, nationalities, and religious or political groups. `Art/Media' covers events, works of art, and products. Finally, `Etcs' comprises answers that do not fit into the other categories.

\section{Additional Related Works}\label{sec:additional_related}
\paragraph{Continual Learning}
Continual learning (CL) is often categorized in three directions: \emph{Regularization-based} approaches\,\citep{LWF_ECCV_16, lee2017overcoming, yoon2023continual} aim to regularize the changes of model parameters to avoid forgetting previous knowledge during continual learning; \emph{Architecture-based} approaches\,\citep{rusu2016progressive, mallya2018piggyback, hung2019increasingly, kang2022soft} utilize different parameters or modules for each task to prevent forgetting; and \emph{Replay-based} approaches\,\citep{rebuffi2017icarl, shin2017continual, rolnick2019experience} store a subset of training samples or other useful data in a replay buffer and learn new tasks by referring to the buffer. 

Along with the remarkable advances in vision-based continual learning, the importance of continual learning for language models has been recognized in recent days~\citep{chen2020recall,  qin2022elle,dhingra2022time,razdaibiedina2023progressive,chen2023lifelong,cole2023salient}. However, most of these works focus on domain-incremental CL, which continually learn different domain corpora such as bio-medical papers to physics papers~\citep{jin2021lifelong, qin2022elle}, or task-incremental CL~\cite{chen2020recall, razdaibiedina2023progressive}.
However, research on temporal evolving continual learning is yet under-explored. 

\paragraph{Model Editing}
Model editing is proposed to keep language models up-to-date and fix any errors in their existing knowledge. There are four main approaches for model editing. 
Memory-based approaches~\cite{mitchell2022memory, zhong2023mquake, zheng2023can} retrieve the most relevant edit facts from external memory. Parameter-expansion approaches~\cite{huang2023transformer} train additional parameters with modified knowledge.
Locate-then-edit approaches~\cite{meng2022locating, meng2022mass} identify the specific parts of the model that need changes and updates them directly. Meta-learning based approaches~\cite{de2021editing, mitchell2021fast} employ a hyper-network trained to predict the necessary gradient update for editing. However, model editing studies overlook multiple updates scenario (i.e., more than 2 update steps), or focus only on knowledge update, disregarding knowledge addition. Moreover, they update knowledge in fine-tuning stage, but continual learning learns and update knowledge during continual pre-training, which enables large amount of knowledge update and close to real-world scenario.